\documentclass[letterpaper, 10 pt, conference]{ieeeconf}  
\IEEEoverridecommandlockouts                              
\usepackage{graphics} 
\usepackage{mathptmx} 
\usepackage{times} 
\usepackage{amsmath, bm} 
\usepackage{amssymb}  
\usepackage{amsfonts}
\usepackage{commath}
\usepackage{lipsum}
\usepackage{tikz}
\usepackage{algorithm}
\usepackage{algpseudocode}
\usepackage[noabbrev]{cleveref}
\usepackage{interval}
\usepackage[dvipsnames]{xcolor}
\usepackage{booktabs, makecell, tabularx, multirow}
\usepackage{optidef}
\usepackage{ifthen}
\usepackage[separate-uncertainty = true,multi-part-units=single]{siunitx}
\DeclareMathOperator*{\argmax}{arg\,max}
\DeclareSIUnit\mmHg{mmHg}
\DeclareSIUnit\BPM{BPM}

\usepackage{cite}

\newboolean{anonymous}
\setboolean{anonymous}{false}

\title{\LARGE \bf Model Predictive Control with Reference Learning for Soft Robotic Intracranial Pressure Waveform Modulation}
\ifthenelse{\boolean{anonymous}}
{
\author{Anonymous Authors%
\thanks{Anonymous authors' affiliation.}
\thanks{Anonymous authors' funding.}
\thanks{Anonymous authors' e-mail.}}
}
{
\author{Fabian Flürenbrock, Yanick Büchel, Johannes Köhler, Marianne Schmid Daners, and Melanie N. Zeilinger%
\thanks{This work was supported by the Swiss National Science Foundation (Grant 315230 184913) and the NCCR Automation (51NF40 180545). All authors are with the Institute for Dynamic Systems and Control at ETH Zurich, Switzerland. E-mail: {\tt\small \{ffluerenb, ybuechel, jkoehle, marischm, mzeilinger\}@ethz.ch}}} 
}

\begin{document}
\maketitle
\thispagestyle{empty}
\pagestyle{empty}

\begin{abstract}
This paper introduces a learning-based control framework for a soft robotic actuator system designed to modulate intracranial pressure (ICP) waveforms, which is essential for studying cerebrospinal fluid dynamics and pathological processes underlying neurological disorders.
A two-layer framework is proposed to safely achieve a desired ICP waveform modulation. 
First, a model predictive controller (MPC) with a disturbance observer is used for offset-free tracking of the system's motor position reference trajectory under safety constraints.
Second, to address the unknown nonlinear dependence of the ICP on the motor position, we employ a Bayesian optimization (BO) algorithm used for online learning of a motor position reference trajectory that yields the desired ICP modulation.
The framework is experimentally validated using a test bench with a brain phantom that replicates realistic ICP dynamics in vitro. 
Compared to a previously employed proportional-integral-derivative controller, the MPC reduces mean and maximum motor position reference tracking errors by \qty{83}{\percent} and \qty{73}{\percent}, respectively. 
In less than 20 iterations, the BO algorithm learns a motor position reference trajectory that yields an ICP waveform with the desired mean and amplitude. 
\end{abstract}

\section{INTRODUCTION}\label{sec:introduction}
The regulation of pressure is a fundamental requirement in many biological systems.
Specifically, intracranial pressure (ICP) plays a crucial role in the brain and central nervous system. 
It affects cerebrospinal fluid (CSF) dynamics and also directly impacts neurological health.
Hydrocephalus is a medical condition characterized by disturbed CSF dynamics and an excessive accumulation of CSF within the cerebral ventricles~\cite{Rekate2008}, commonly treated using shunt systems~\cite{Fluerenbrock2025VIEshunt}.
While mean ICP is an important clinical marker for this condition, the pulsatile components of the ICP waveform have gained increasing attention in research as these may reflect and contribute to the underlying pathological processes~\cite{Evensen2020, Green2021}. 
Monitoring, understanding, and controlling the ICP waveform is therefore essential for advancing the study and therapy of neurological disorders such as hydrocephalus.

Various research groups have developed robotic systems to perform active modulation of the ICP waveform in animal studies~\cite{Mancinelli1976, DiRocco1977, DiRocco1978, Pettorossi1978, Luciano2017, Luciano2020, Qvarlander2023, Doron2020, Doron2021, Doron2022, Fluerenbrock2024TBME}. 
While the respective research objectives slightly differ among them, the developed systems are commonly based on motorized pumps, which cyclically inflate and deflate a balloon that is placed within a location of the CSF system such as the cerebral ventricles. 
As the cardiac cycle is the primary driver of the pulsatile ICP waveform~\cite{Linninger2016}, inflation and deflation of the balloon are gated to the electrocardiogram (ECG) and performed during the systolic phase of the heartbeat, during which ventricular blood ejection raises arterial pressure and drives ICP pulsation.
The control of these motorized pump systems is typically implemented using open-loop or proportional-integral-derivative (PID) controllers.
However, since the exact influence of the motor position on the ICP is unknown, achieving the desired ICP waveform modulation with these control approaches requires \emph{time-consuming and error-prone hand-tuning} of the motor's reference trajectory and control parameters.
An additional challenge is that the safety of the aspired in vivo intervention must be ensured during the whole tuning process and throughout the subsequent deployment. 
Incorrect inflation and deflation of the implanted balloon can not only cause undesired baseline shifts in the ICP but also lead to pathological levels of ICP or even mechanical failure of the balloon.
Accurate and offset-free tracking of the motor position reference trajectory and the integration of safety constraints are thus required.
Since the systolic phase of the heartbeat is only about \qty{350}{\ms} long~\cite{Fridericia2003}, this task poses a challenging control problem for the soft robotic setup.

\textit{Contribution:} This paper introduces a learning-based control framework to improve both the safety and performance of a soft robotic actuator system~\cite{Fluerenbrock2024TBME} used for ICP waveform modulation in hydrocephalus research.
To this end, three contributions are presented: 
\begin{enumerate}
    \item A model predictive controller (MPC)~\cite{Rawlings2017MPCBook} for reference tracking is designed and augmented with a disturbance observer that exploits the periodic operation of the system. The MPC scheme improves system performance through offset-free tracking of the motor position reference trajectory and enhances safety through explicit integration of the system's state and input constraints.
    \item A Bayesian optimization (BO)~\cite{Shahriari2016} algorithm for data-efficient and automated MPC reference learning is designed. The BO algorithm improves system performance by iteratively tuning the parameterized motor position reference trajectory of the MPC during run-time, thereby learning the optimal reference trajectory that results in the desired ICP waveform modulation.
    \item The introduced learning-based control framework is deployed on the soft robotic actuator system~\cite{Fluerenbrock2024TBME}, and the active ICP waveform modulation is experimentally investigated using a mechatronic test bench that features a brain phantom capable of replicating physiological ICP waveforms in vitro~\cite{Castelar2022}.
\end{enumerate}

\begin{figure*}
    \centering
    \includegraphics[width=\textwidth]{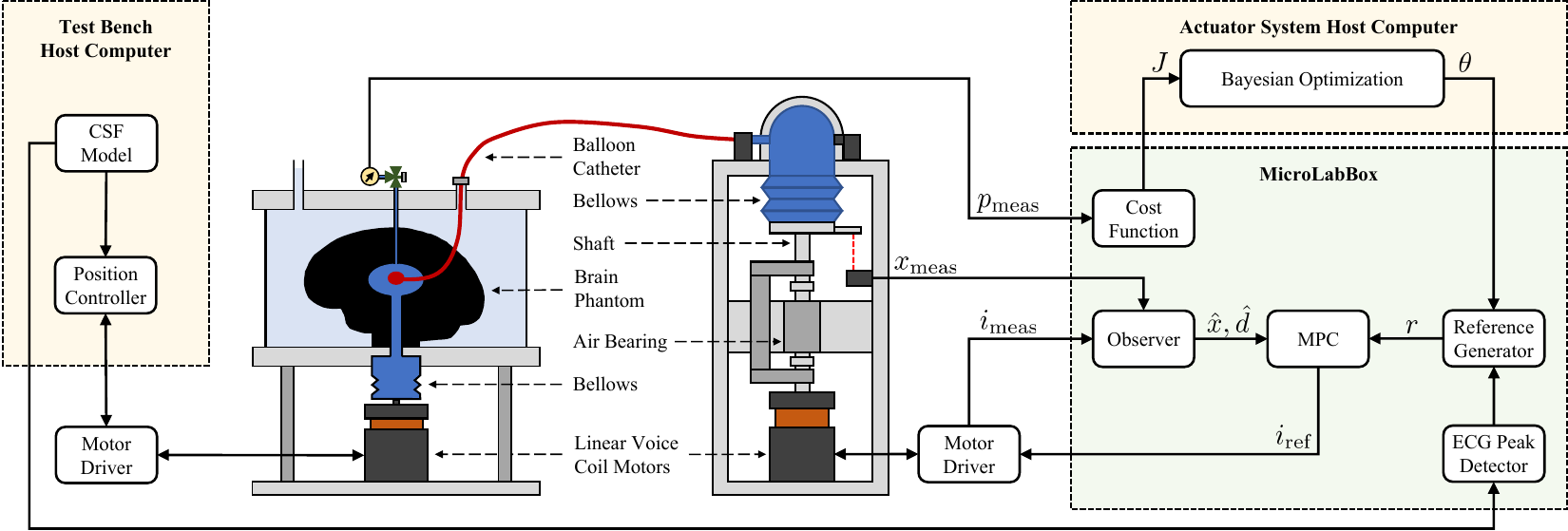}
    \caption{Schematic overview of the experimental hardware setup. The soft robotic actuator system has been developed in~\cite{Fluerenbrock2024TBME} and is used for intracranial pressure (ICP) modulation. The mechatronic test bench has been developed in~\cite{Castelar2022} and is used for the physical replication of ICP waveforms. The balloon catheter of the soft robotic actuator system is inserted into the brain phantom of the test bench to enable ICP waveform modulation. The pressure measurements of the brain phantom and the electrocardiogram (ECG) of the cerebrospinal fluid (CSF) model used for simulating physiological ICP waveforms are fed back to the soft robotic actuator system.}
    \label{fig:setup}
\end{figure*}

\ifthenelse{\boolean{anonymous}}
{
\newpage
}
{
}

\textit{Outline:} In Section~\ref{sec:problem}, we abstract the problem considered in this work into a mathematical description and highlight the key technical difficulties.
In Section~\ref{sec:methods}, we provide a general methodology to address the considered problem based on an MPC formulation for reference tracking and a BO algorithm for reference learning.
In Section~\ref{sec:experiments}, we discuss the deployment and experimental validation of the methodology on hardware.
Section~\ref{sec:conclusion} concludes the paper.

\section{PROBLEM FORMULATION}\label{sec:problem}
The experimental hardware setup of this work, composed of the soft robotic actuator system~\cite{Fluerenbrock2024TBME} and the mechatronic test bench~\cite{Castelar2022}, is sketched in Figure~\ref{fig:setup}.
To describe the soft robotic actuator system, the following discrete-time linear time-invariant system is considered
\begin{subequations} \label{eq:system_model}
\begin{align}
    x(k\!+\!1) &= A x(k) + B u(k) +B_\mathrm{d} d(k),\\
    y(k) &= C x(k) + C_\mathrm{d} d(k),
\end{align}
\end{subequations}
where $k \in \mathbb{N}$ denotes the discrete time step, the system state $x \in \mathbb{R}^2$ consists of the motor position and velocity, the input $u \in \mathbb{R}$ is the applied motor current, the output $y \in \mathbb{R}$ is the measured motor position, and the additive disturbance $d \in \mathbb{R}$ captures all unmodeled effects. 
We consider polytopic state and input constraints that are denoted as $\mathbb{X} \subset \mathbb{R}^2$ and $\mathbb{U} \subset \mathbb{R}$, respectively.
We assume that the system matrix $A$ and input matrix $B$ are approximately known through system identification.
We further consider a reference $r \in \mathbb{R}$ that is to be tracked by the system output.
The shape of the reference trajectory for a full inflation and deflation cycle of the balloon is shown in Figure~\ref{fig:reference}.
It is parameterized by a set of variables, among which delay and magnitude are considered the most influential.
These two are therefore selected for tuning and jointly denoted by $\theta \in \mathbb{R}^2$.
As can be seen in Figure~\ref{fig:setup}, the motor position $y$ affects the inflation of the implanted balloon.
This in turn affects the ICP, which is denoted by $p \in \mathbb{R}$.
Given the periodicity of the cardiac cycle, which is assumed to have a constant period length $T \in \mathbb{N}$, the relationship between motor position and ICP can be characterized as
\begin{equation} \label{eq:unknown_mapping}
    \mathbf{p} = \Delta ( \mathbf{y} ),
\end{equation}
where $\Delta$ is an unknown function that maps the output sequence $\mathbf{y} = \{ y_0, \ldots, y_{T-1}\}$ of the soft robotic actuator system to the ICP sequence $\mathbf{p} = \{ p_0, \ldots, p_{T-1}\}$, both being of period length $T$.
Due to the highly nonlinear pressure-volume dynamics of the implanted balloon and the unpredictable physiological disturbances during in vivo application, the mapping described in~\eqref{eq:unknown_mapping} is difficult to model accurately.

The overarching goal of this work is to safely achieve a desired ICP waveform, which is defined through the mean pressure $p_\text{mean} = \text{mean}(\mathbf{p})$ and the peak-to-peak amplitude $p_\text{amp} = \text{max}(\mathbf{p}) - \text{min}(\mathbf{p})$ of the ICP sequence $\mathbf{p}$ during a single cardiac cycle.
For the specific example of replicating normal pressure hydrocephalus in vivo, as considered in Section~\ref{sec:experiments}, the peak-to-peak amplitude is to be increased to a reference amplitude $p_\text{amp}^\text{ref}$ without raising the mean ICP from its baseline level $p_\text{mean}^\text{base}$ at times when no actuation is performed.
To achieve the desired ICP waveform modulation with the soft robotic actuator system, the problem is decomposed into two independent sub-problems, namely the reference tracking problem and the reference learning problem. The respective objectives of these two sub-problems are defined in detail as follows:
\begin{enumerate}
    \item \textit{Reference tracking:} Design a controller that steers the system output $y$ as close as possible to the reference trajectory $\mathbf{r}(\theta) = \{ r_0, \ldots, r_{T-1}\}$ while compensating for the model mismatch $d$ and satisfying the system constraints $x \in \mathbb{X}$ and $u \in \mathbb{U}$.
    \item \textit{Reference learning:} Find the optimal parameters $\theta^\star$ from a compact set of feasible parameters $\Theta \subset \mathbb{R}^2$, which yield a reference trajectory $\mathbf{r}(\theta^\star)$ that achieves the desired ICP waveform in terms of $p_\text{amp}^\text{ref}$ and $p_\text{mean}^\text{base}$. \\
\end{enumerate}

\section{METHODS}\label{sec:methods}
This section outlines the proposed methods for the soft robotic actuator system to safely achieve a desired ICP waveform modulation.
Section~\ref{sec:methods_control} presents the MPC with disturbance observer for accurate and safe tracking of arbitrary motor reference trajectories.
Section~\ref{sec:methods_learning} presents the BO approach used for learning an optimal motor reference trajectory that yields the desired waveform modulation.
The full learning-based control framework for the soft robotic actuator system is summarized in Algorithm~\ref{alg:chapter4}.

\subsection{Reference Tracking}\label{sec:methods_control}
This work employs an output reference tracking MPC that uses an augmented linear model with disturbance observer to achieve offset-free reference tracking~\cite{Muske2002, Pannocchia2003, Maeder2009}.
The disturbance observer and MPC formulation are described in the following in detail.

\textit{Disturbance observer:} To account for the model mismatch between the identified linear system model in~\eqref{eq:system_model} and the true system, the linear system model is augmented with the disturbance state $d$, which results in the following augmented system description
\begin{subequations} \label{eq:augmented_model}
    \begin{align} 
        \begin{bmatrix} x(k\!+\!1) \\ d(k\!+\!1) \end{bmatrix}
        &= 
        \underbrace{\begin{bmatrix} A & B_\mathrm{d} \\ 0 & I \end{bmatrix}}_{A_\mathrm{a}}
        \begin{bmatrix} x(k) \\ d(k) \end{bmatrix}
        + 
        \underbrace{\begin{bmatrix} B \\ 0 \end{bmatrix}}_{B_\mathrm{a}} u(k), \\
        y(k) &= 
        \underbrace{\begin{bmatrix} C & C_\mathrm{d} \end{bmatrix}}_{C_\mathrm{a}}
        \begin{bmatrix} x(k) \\ d(k) \end{bmatrix},
    \end{align}
\end{subequations}
where $I$ describes an identity matrix of appropriate size, and $A_\mathrm{a}$, $B_\mathrm{a}$, and $C_\mathrm{a}$ are the augmented state, input, and output matrix, respectively.
For online estimation of the augmented state vector, the following Luenberger observer \cite{Luenberger1966, Pannocchia2015a} is designed
\begin{subequations} \label{eq:observer}
    \begin{align}
        \begin{bmatrix} \hat{x}(k\!+\!1) \\ \hat{d}(k\!+\!1) \end{bmatrix}
        &= A_\mathrm{a} \begin{bmatrix} \hat{x}(k) \\ \hat{d}(k) \end{bmatrix}
        + B_\mathrm{a} u(k) + \underbrace{\begin{bmatrix} L_\mathrm{x} \\ L_\mathrm{d} \end{bmatrix}}_L \left( y(k) - \hat{y}(k) \right), \\
        \hat{y}(k) &= C_\mathrm{a} \begin{bmatrix} \hat{x}(k) \\ \hat{d}(k) \end{bmatrix},
    \end{align}
\end{subequations}
where the observer gain $L$ is chosen via pole placement such that $\Tilde{A} - L \Tilde{C}$ is Schur stable. 
The estimated disturbances of the full pulse sequence of length $T$ are saved in the disturbance memory sequence $\hat{\mathbf{d}} = \{ \hat{d}_0, \ldots, \hat{d}_{T-1}\}$ to be used by the MPC in the next period.

\textit{Model predictive control:} To steer the system described in~\eqref{eq:system_model} toward its reference trajectory, an output reference tracking MPC with prediction horizon $N$ is used.
Given the reference trajectory $\mathbf{r}(\theta) = \{ r_0, \ldots, r_{T-1}\}$, the estimated disturbance sequence of the previous period $\hat{\mathbf{d}}$ and the current state estimate $\hat{x}(k)$ obtained from the observer in~\eqref{eq:observer}, the MPC controller $\pi_\text{MPC}$ computes at each time step $k$ the input $u(k)$ by solving the following optimization problem
\begin{subequations} \label{eq:MPC}
\begin{align}
        \min_{\mathbf{x}, \mathbf{u}} \quad & \sum_{i=0}^{N-1} \lVert C x_i + C_\mathrm{d} \hat{d}(k+i-T) - r_{k+i} \rVert_2^2, \qquad \qquad \\
        \textrm{s.t. } \:\; & \forall i = 0, \ldots, N\!-\!1, \notag \\ 
        & x_{i+1} = A x_{i} + B u_{i} + B_\mathrm{d} \hat{d}(k+i-T), \\
        & (x_i, u_i) \in \mathbb{X} \times \mathbb{U}, \\
        & x_0 = \hat{x}(k).
\end{align}
\end{subequations}
In contrast to standard offset-free MPC formulations~\cite{Muske2002, Pannocchia2003, Maeder2009}, we leverage the iterative nature of the problem by utilizing the disturbance estimates from the previous pulse sequence in the MPC prediction, similar to~\cite{Pabon2024}. 
The considered singular output cost is applicable since the model (A,B,C) is flat (otherwise an input regularization should be added, see~\cite{Kohler2024Analysis, Köhler2022a}).
Closed-loop stability can be ensured if the reference trajectory $\mathbf{r}(\theta)$ is feasible and a sufficiently long prediction horizon $N$ is utilized~\cite{Köhler2022a, Kohler2024Analysis}.

\begin{figure}
\centering
\begin{tikzpicture}
    \draw[very thick, lightgray] (0,1)--(0.75,1);
    \draw[very thick, RoyalBlue] (0.75,1)--(1.75,1)--(2.75,3)--(3.75,3)--(4.75,1)--(5,0.5)--(5.4,1)--(6,1);
    \draw[very thick, lightgray] (6,1)--(7.25,1);
    \draw[thick, ->] (0,0)--(7.5,0) node[midway, below, yshift=-0.1cm] {\footnotesize Time [s]};
    \draw[thick, ->] (0,0)--(0,3.75) node[midway, left, rotate=90, yshift=0.3cm, xshift=1.05cm] {\footnotesize Position [mm]}; 
    \draw[thick, OliveGreen, ->] (0.75,1.55)--(0.75,1.05) node[yshift=0.75cm]{\scriptsize trigger};
    \draw[thick, Black, <->] (5.5,0.95)--(5.5,0.5) node[right, midway]{\scriptsize neg. magnitude};
    \draw[thick, BrickRed, <->] (2.9,2.9)--(2.9,1.1) node[right, midway]{\scriptsize magnitude};
    \draw[thick, BrickRed, <->] (0.75,0.8)--(1.75,0.8) node[below, midway]{\scriptsize delay};
    \draw[thick, Black, <->] (0.75,0.2)--(6,0.2) node[above, midway]{\scriptsize total length};
    \draw[thick, Black, <->] (1.775,3.2)--(2.725,3.2) node[above, midway]{\scriptsize rise time};
    \draw[thick, Black, <->] (2.775,3.2)--(3.725,3.2) node[above, midway, yshift=-0.03cm]{\scriptsize up time};
    \draw[thick, Black, <->] (3.775,3.2)--(4.725,3.2) node[above, midway]{\scriptsize fall time};
    \draw[thick, Black, <->] (4.775,1.2)--(5.375,1.2) node[above, midway]{\scriptsize \shortstack{down\\time}};
    \draw[thick, dashed, lightgray] (1.85,1)--(5.3,1);
    \draw[thick, dashed, lightgray] (5.025,0.45)--(5.55,0.45);
    \draw[thick, dashdotted, lightgray] (1.75,3.25)--(1.75,1.05);
    \draw[thick, dashdotted, lightgray] (2.75,3.25)--(2.75,3.05);
    \draw[thick, dashdotted, lightgray] (3.75,3.25)--(3.75,3.05);
    \draw[thick, dashdotted, lightgray] (4.75,3.25)--(4.75,1.05);
\end{tikzpicture}
\caption{Visualization of the periodically triggered and pulse-shaped reference trajectory for the motor position of the soft robotic actuator system. The delay and magnitude (red arrows) are tunable parameters that are determined via Bayesian optimization. The next pulse can only be triggered by the detection of an QRS complex in the electrocardiogram after the previous pulse has been completed. In between pulses, the motor is regulated to its baseline position (gray line).}
\label{fig:reference}
\end{figure}
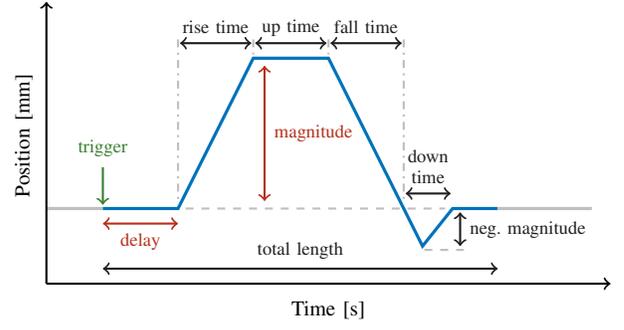

\begin{algorithm}
\caption{Offset-free MPC with Reference Learning}
\label{alg:chapter4}
\begin{algorithmic}[1]
\State \textbf{Input:} Number of max iterations $n_\mathrm{m}$, number of random samples $n_\mathrm{r}$, number of evaluation periods per iteration $n_\mathrm{p}$, number of time steps per period $T$, hyperparameter $\beta$, disturbance threshold $\epsilon$, and feasible parameter set $\Theta$
\State \textbf{Initialize:} GP prior $J(\theta)\sim GP(\mu_0,\sigma_0)$, dataset $D_0$, and sequence of disturbance estimates $\hat{\mathbf{d}}$
\For{$n=1, \ldots, n_\mathrm{m}$} 
    \If{$n \leq n_\mathrm{r}$}
        \State $\theta_n \gets$ Sample parameters randomly from $\Theta$
    \Else
        \State $\mu_{n-1}(\theta), \sigma_{n-1}(\theta) \gets$ Update GP based on $D_{1:n-1}$
        \State $\theta_n \gets \argmax_{\theta \in \Theta} \; \mu_{n-1}(\theta) + \beta \cdot \sigma_{n-1}(\theta)$
    \EndIf
    \State $l \gets 0$ Initialize evaluation period counter
    \While{$l \leq n_\mathrm{p}$}
        \For{$k=0, \ldots, T-1$}
            \State $y_k \gets$ Take measurement
            \State $\hat{x}_k, \hat{d}'_k \gets$ Estimate state and disturbance
            \State $u_k \gets$ Compute input using $\pi_\text{MPC}(\hat{x}_k, \mathbf{\hat{d}}, \mathbf{r}(\theta_n))$
        \EndFor
        \If{$\mid \hat{d}'_k - \hat{d}_k\mid \leq \epsilon \; \forall k \in [0,T\!-\!1]$}
            \State $J_n^l \gets$ Evaluate parameter cost of period $l$
            \State $l \gets l + 1$ Advance counter
        \EndIf
        \State $\hat{\mathbf{d}} \gets \hat{\mathbf{d}}'$ Update disturbance estimates
    \EndWhile
    \State $J_n \gets$ Evaluate cost for $\theta_n$ as average of $n_\mathrm{p}$ periods
    \State $D_{0:n} \gets$ Update dataset with $D_{0:n-1} \cup \{(\theta_n, J_n)\}$  
\EndFor
\State \textbf{Return:} Optimal parameters $\theta^*$
\end{algorithmic}
\end{algorithm} 

\subsection{Reference Learning} \label{sec:methods_learning}
To achieve the desired ICP waveform modulation, the correct reference trajectory $\mathbf{r}(\theta)$ for the MPC must be found.
Since the mapping between the system's output sequence $\mathbf{y}$ and the ICP sequence $\mathbf{p}$ in~\eqref{eq:unknown_mapping} is unknown, a black-box optimization approach based on BO is chosen to learn the optimal motor position reference trajectory.
BO has been widely applied in control, particularly for automatic controller tuning~\cite{Berkenkamp2016d, Marco2016, Sorourifar2021}, motivating its use in this work for automatic tuning of the reference trajectory parameterization shown in Figure~\ref{fig:reference}.
Specifically, the aim of the proposed BO algorithm is to find the optimal parameter values $\theta^\star$ that minimize a cost function $J(\theta)$, which quantifies the deviation between the observed and desired ICP waveform characteristics. 
The cost function is evaluated over the full period of length $T$ and defined as
\begin{equation}\label{eq:bo_cost}
    J(\theta) = - \left( p_\text{amp} - p_\text{amp}^\text{ref} \right)^2 - \lambda \sum_{i=0}^{T-1} p_i^2.
\end{equation}
The first term penalizes the difference between the observed and desired peak-to-peak amplitude of the ICP, whereas the second term penalizes the energy of the ICP signal in order to prevent any increase in the mean ICP from the baseline value. 
The hyperparameter $\lambda \geq 0$ trades off the two terms in the cost function.
To reduce the variance in the cost evaluation, the parameter cost is computed as the average over a series of $n_\mathrm{p}$ measured periods.
To filter out transient effects in the reference tracking performance caused by changes in the parameter setting $\theta$, measured periods are counted toward these $n_\mathrm{p}$ periods only after the difference between two consecutive estimates of the disturbance sequence $\hat{\mathbf{d}}$ falls below a predefined threshold $\epsilon \in \mathbb{R}_{>0}$.

In the proposed BO algorithm, a Gaussian process (GP)~\cite{Rasmussen2006} with a squared-exponential kernel is employed to construct a probabilistic surrogate model of the cost function~\eqref{eq:bo_cost}. 
The algorithm is initialized by uniformly sampling and evaluating $n_\mathrm{r}$ parameters from the feasible parameter set $\Theta$. 
In each of the subsequent iterations, indexed by $n$, a new candidate $\theta_n$ is selected by maximizing an acquisition function based on the upper confidence bound~\cite{Srinivas2010}:
\begin{equation} \label{eq:bo_acq_fun} 
    \theta_{n} = \argmax_{\theta \in \Theta} \; \mu_{n-1}(\theta) + \beta \cdot \sigma_{n-1}(\theta),
\end{equation} 
where $\mu_{n-1}(\theta)$ and $\sigma_{n-1}(\theta)$ denote the posterior mean and standard deviation of the GP model at iteration $n\!-\!1$, conditioned on the dataset $D_{1:n-1} = \{(\theta_1, J_1), \ldots, (\theta_{n-1}, J_{n-1})\}$ of previous experiments and their corresponding costs. 
The hyperparameter $\beta>0$ regulates the trade-off between exploitation and exploration.

\section{EXPERIMENTS}\label{sec:experiments}
This section presents the experimental evaluation of the proposed learning-based control framework for the soft robotic actuator system.
While Section~\ref{sec:application_tracking} evaluates the MPC tracking performance, Section~\ref{sec:application_modulation} evaluates the full framework's ICP waveform modulation performance.
In both experiments, commercially available Fogarty balloon catheters with a maximum volume of \qty{2.5}{\milli\liter} were used as the soft robotic end actuators.

\subsection{Reference Tracking}\label{sec:application_tracking}
Accurate tracking of the actuator system's motor reference trajectory is essential for enabling precise ICP waveform modulation. 
Deviations from the desired trajectory can introduce unintended pressure fluctuations, potentially compromising the system’s effectiveness. 
The first experiment was therefore dedicated to the evaluation of the motor position tracking performance of the proposed MPC scheme. 
To this end, the soft robotic actuator system shown in Figure~\ref{fig:setup} was first evaluated in standalone operation, i.e., without integration into the test bench.
During the experiment, the system was tasked with following a reference trajectory similar to those reference trajectories used during the later application for ICP waveform modulation shown in Figure~\ref{fig:reference}. 
The motor position was initialized at a baseline of \qty{0.63}{\mm}, just before the balloon inflation begins. 
The reference trajectory consisted of pulses with a magnitude of \qty{0.5}{\mm} at \qty{60} beats per minute (BPM). 
Three control approaches were compared: (i)~baseline PID controller, (ii)~MPC without disturbance observer, and (iii)~MPC with disturbance observer (offset-free MPC).
For the MPC-based controllers, the system matrix $A$ and input matrix $B$ were identified from representative experimental data using the prediction error method \cite{Ljung1999}. 
Both MPC schemes used a prediction horizon of $N = 15$. 
For the Luenberger observer, the disturbance matrices were chosen as $B_\mathrm{d} = 0$ and $C_\mathrm{d} = 1$, ensuring observability of the augmented system~\eqref{eq:augmented_model}. 
The system's state constraints $\mathbb{X}$ and input constraints $\mathbb{U}$ are provided in Table~\ref{tab:methods_mpc_constraints}. 
The disturbance observer and MPC are implemented on a real-time embedded computer (MicroLabBox, dSpace, Paderborn, Germany) with Acados~\cite{Verschueren2022} and run at a frequency of \qty{100}{\Hz}.

\begin{figure*}[!t]
    \centering
    \includegraphics[width=\textwidth]{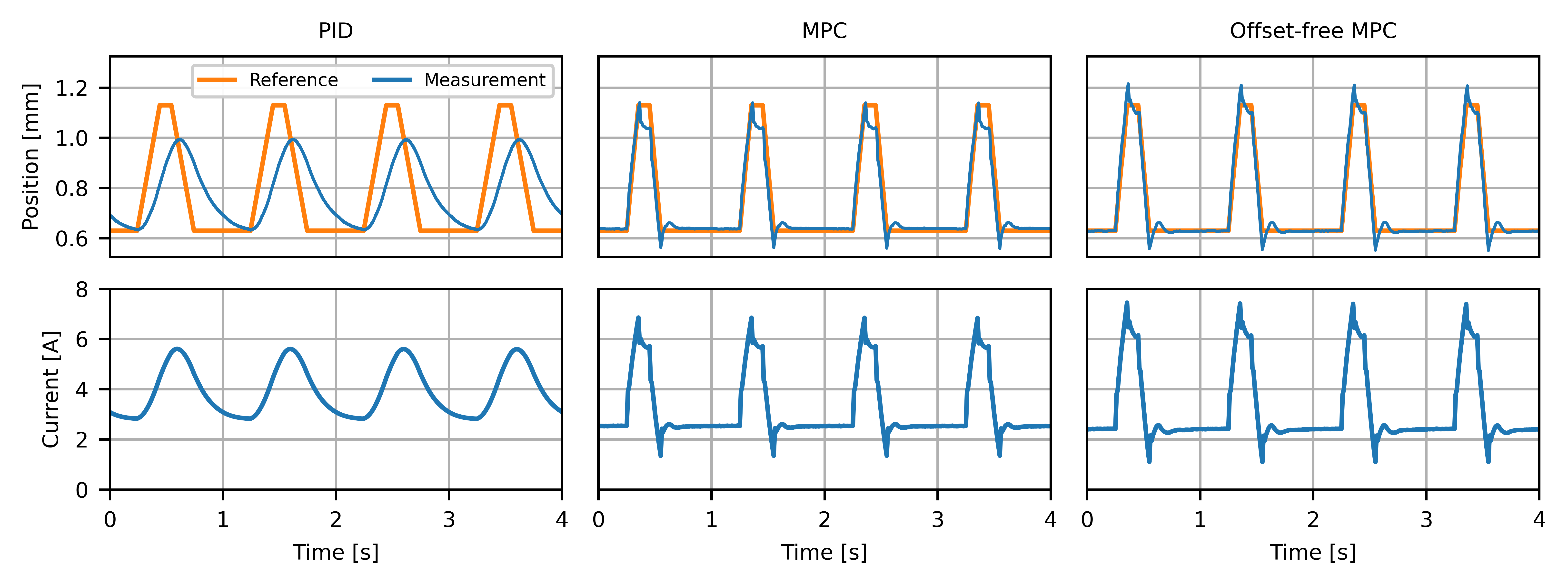} 
    \caption{Reference tracking experiment: Comparison of the baseline proportional-integral-derivative (PID) controller, the model predictive controller (MPC) without disturbance observer, and the MPC with disturbance observer (offset-free MPC). The upper panels show the motor position, whereas the lower panels show the motor current. Controllers were tested with the same \qty{60}{\BPM} reference trajectory, similar to Figure~\ref{fig:reference} but without negative magnitudes.}
    \label{fig:tracking}
\end{figure*} 
\begin{table}[!t]
    \centering
    \setlength\extrarowheight{2pt}
    \caption{State and Input Constraints of the System}
    \label{tab:methods_mpc_constraints}
        \begin{tabularx}{\columnwidth}{>{\hsize=1.0\hsize}X>{\hsize=0.8\hsize}X>{\hsize=0.8\hsize}X>{\hsize=0.5\hsize}X}
        \toprule
        \textbf{Constraint \quad\quad} & \textbf{Lower Limit \quad} & \textbf{Upper Limit \quad} & \textbf{Unit} \\
        \midrule
        \textbf{Motor Position} & 0 & 2.6 & \qty{}{\mm} \\
        \textbf{Motor Velocity} & -100 & +100 & \qty{}{\mm\per\s} \\
        \textbf{Motor Input} & -10 & +10 & \qty{}{\ampere} \\
        \bottomrule
    \end{tabularx}
\end{table}

The tracking performance of the three compared control approaches is illustrated in Figure~\ref{fig:tracking}. 
The PID controller exhibited significant phase lag and struggled to accurately follow the reference trajectory. 
In contrast, both MPC-based approaches effectively captured the pulse shape since they were able to use the preview of the reference trajectory. 
However, minor undershoots and overshoots following the rising and falling edges of the trajectory were observed. 
Additionally, offsets in the plateau regions were observed when using the MPC without disturbance observer. 
The MPC with disturbance observer achieved offset-free reference tracking, i.e., the controller accurately maintained the desired motor position without steady-state errors during segments with constant motor position reference. 
As PID controller tuning during experiments is time-consuming and error-prone, the baseline PID controller is conservatively tuned to avoid aggressive tracking behavior under changing experimental conditions.
The MPC-based controller, however, allows for explicit constraint handling and transfer to other systems through identification of a simple linear system model.

To quantitatively assess tracking performance, the following two metrics were computed. 
First, the normalized root mean square error (NRMSE) was used to evaluate the average deviation from the reference trajectory while accounting for variations in trajectory magnitude. 
It is defined as 
\begin{equation}
    \mathrm{NRMSE \; [\%]} = 100 * \frac{\sqrt{\frac{1}{M}\sum_{k=0}^{M-1} \left( r_k - y_k \right)^2}}{\mathrm{max}(\mathbf{r}_{M})-\mathrm{min}(\mathbf{r}_{M})},
\end{equation}
where $\mathbf{r}_M = \{ r_0, \ldots, r_{M-1}\}$ and $\mathbf{y}_M = \{ y_0, \ldots, y_{M-1}\}$ represent the reference and system output during the full test sequence of length $M$, respectively. 
Second, the maximum absolute tracking error (MATE) was used to quantify the worst-case deviation between the motor position and its reference trajectory. 
It is formally defined as
\begin{equation}
    \mathrm{MATE \; [mm]} = \mathrm{max}(\mathrm{abs}(\mathbf{r}_{M} - \mathbf{y}_{M})).
\end{equation}
The results for both metrics are summarized in Table~\ref{tab:tracking}. 
Compared to the baseline PID controller, the MPC without a disturbance observer reduced the NRMSE by \qty{75}{\percent} and the MATE by \qty{62}{\percent}.
The augmentation of the MPC with the disturbance observer further enhanced tracking performance.
Compared to the baseline PID controller, the offset-free MPC reduced the NRMSE by \qty{83}{\percent} and the MATE by \qty{73}{\percent}.

\begin{table}[!t]
    \centering
    \setlength\extrarowheight{2pt}
    \caption{Experimental Reference Tracking Results}
    \label{tab:tracking}
    \begin{tabularx}{\columnwidth}{>{\hsize=1\hsize}X S[table-format=2.1] S[table-format=1.3]}
        \toprule
        \textbf{Controller} & {\textbf{NRMSE [\%]\qquad}} & {\textbf{Max. Error [mm]\qquad}} \\
        \midrule
        \textbf{PID}             & 32.3 & 0.316 \\
        \textbf{MPC}             &  8.0 & 0.119 \\
        \textbf{Offset-free MPC} &  5.5 & 0.086 \\
        \bottomrule
    \end{tabularx}
\end{table}

\begin{figure*}[!tbp]
    \centering
    \includegraphics[width=\textwidth]{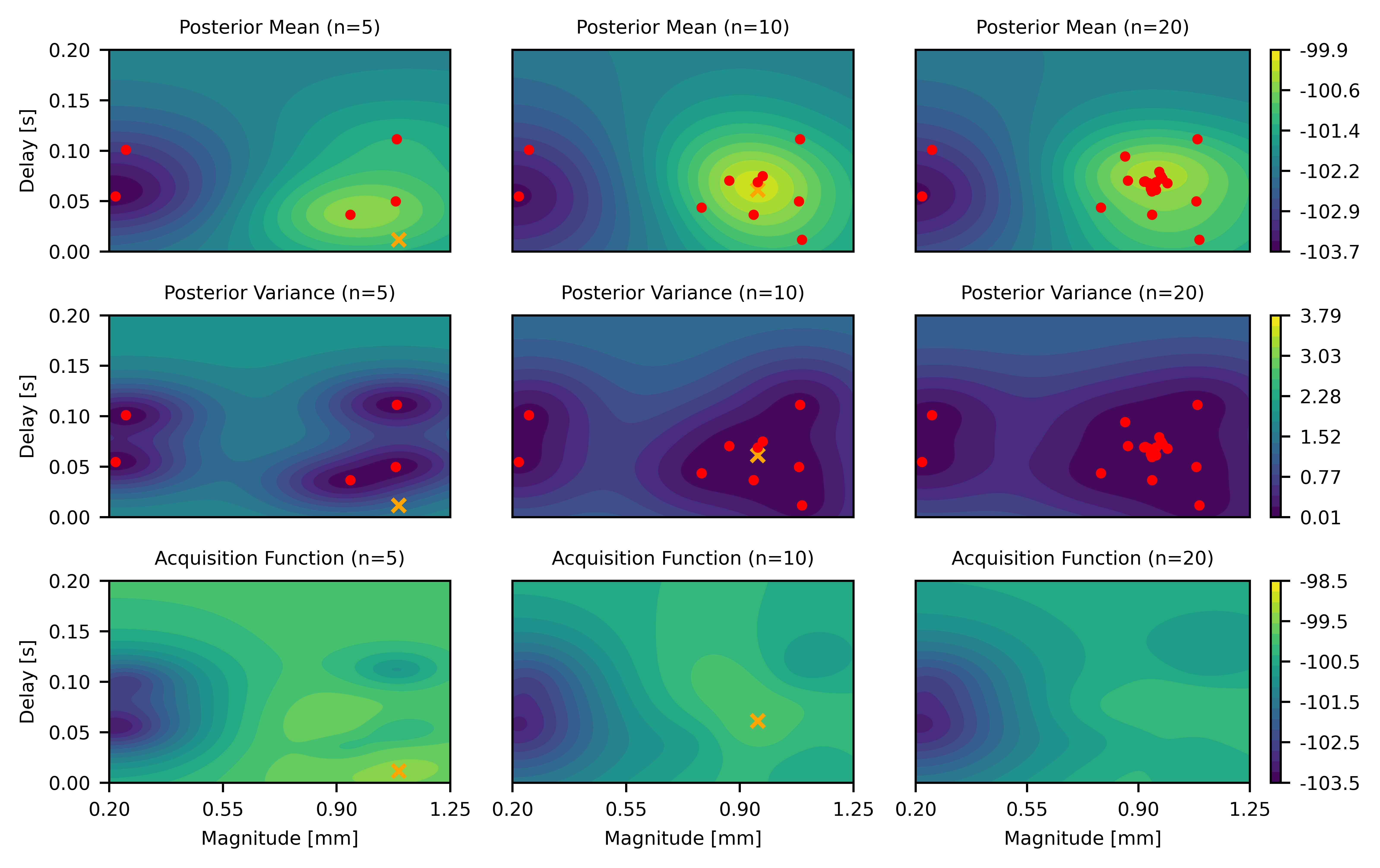}
    \caption{Waveform modulation experiments at \qty{90}{\BPM}: Results of the Bayesian optimization (BO) algorithm used for tuning the delay and magnitude parameter of the controller's motor position reference trajectory. The top panels show the posterior mean of the Gaussian process (GP), the middle panels show the posterior variance of the GP, and the bottom panels show the acquisition function based on the upper confidence bound in~\eqref{eq:bo_acq_fun}. Red dots represent evaluated parameters, whereas orange crosses represent new parameter candidates computed by maximizing the acquisition function. The $n=5$ samples in the leftmost plots represent the randomly sampled parameters used for initialization of the BO algorithm.}
    \label{fig:BO}
\end{figure*}
\begin{figure*}
    \centering
    \includegraphics[width=\textwidth]{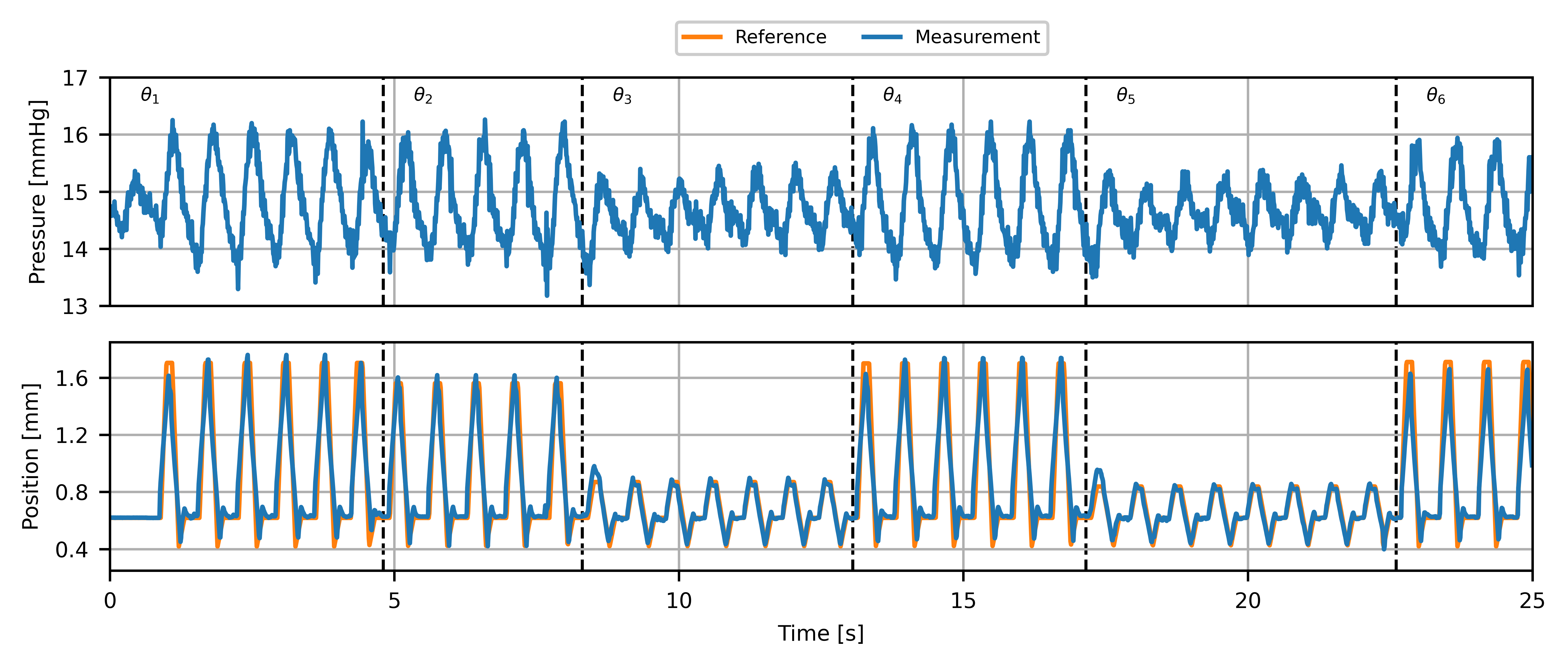}
    \caption{Waveform modulation experiment at \qty{90}{\BPM}: Results of the actuation-based intracranial pressure (ICP) waveform modulation during the first six iterations of the Bayesian optimization (BO) algorithm used for automated reference learning. The upper panel shows the ICP inside the brain phantom of the test bench, whereas the lower panel shows the motor position of the soft robotic actuator system. Changes in the parameterization $\theta$ of the reference trajectory, i.e., BO iterations, are highlighted by vertical dashed lines. A minimum of five cycles were performed per parameter setting.}
    \label{fig:actuation_tuning}
\end{figure*}
\begin{figure}
    \centering
    \includegraphics[width=\columnwidth]{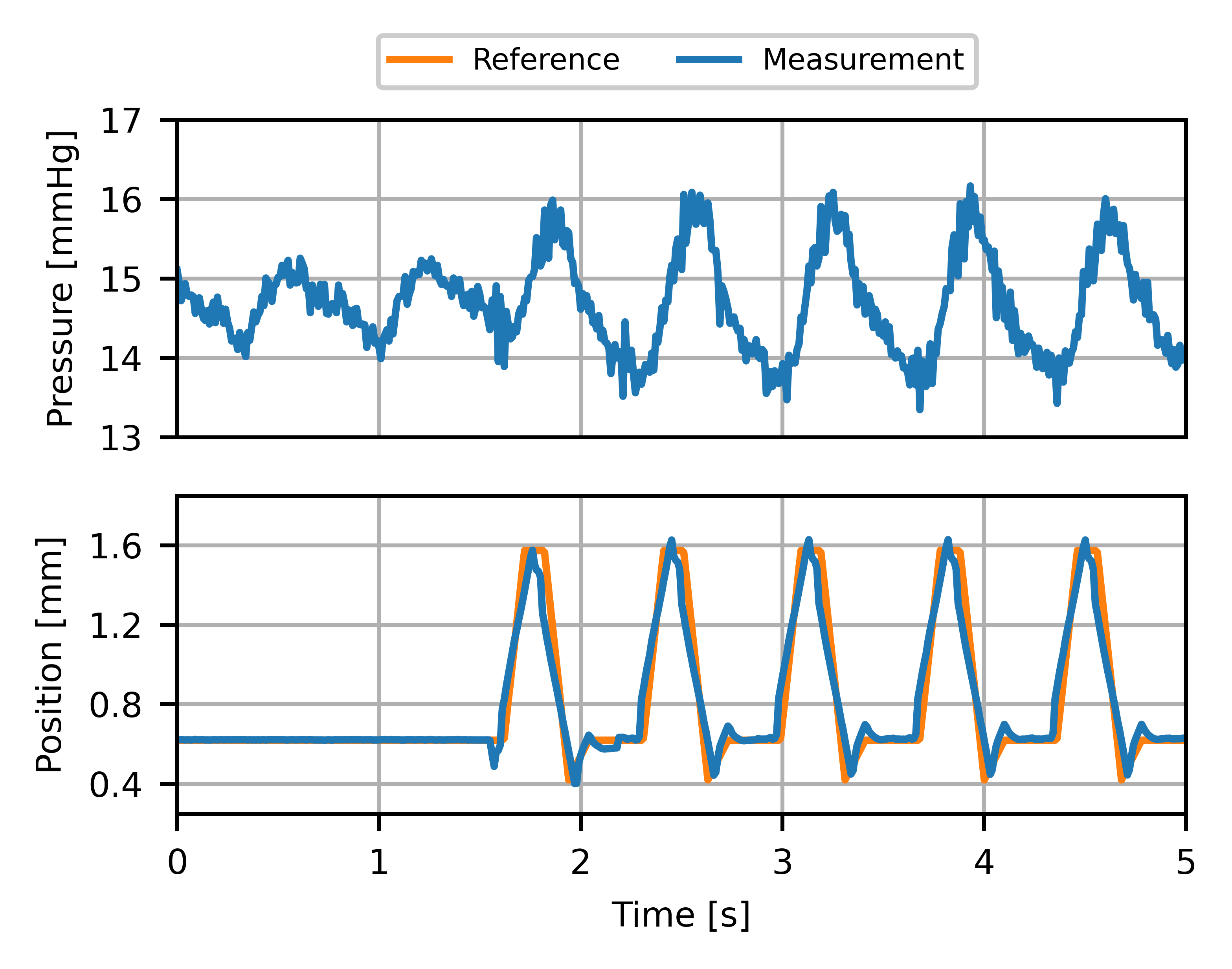}
    \caption{Waveform modulation experiment at \qty{90}{\BPM}: Results of the actuation-based intracranial pressure (ICP) waveform modulation with the final parameter values resulting from the Bayesian optimization algorithm used for automated reference learning. The upper panel shows the ICP inside the brain phantom of the test bench, whereas the lower panel shows the motor position of the soft robotic actuator system. In the beginning, the actuation device is in a steady state and only the natural ICP waveforms as simulated by the test bench are visible in the phantom pressure. In the following, the soft robotic actuator system doubles the peak-to-peak amplitude of the ICP without significantly increasing the mean pressure.}
    \label{fig:actuation_final} \vspace{-2mm}
\end{figure} 

\subsection{Waveform Modulation}\label{sec:application_modulation}
The second experiment was dedicated to evaluating the soft robotic actuator system's ICP waveform modulation capabilities when using the full learning-based control framework.
To this end, the experimental test setup as shown in Figure~\ref{fig:setup} was used and the balloon catheter inserted into the ventricle of the brain phantom.
During this experiment, the motor position reference tracking was fixed to the best-performing reference tracking controller, i.e., the offset-free reference tracking MPC, and BO was utilized to perform reference learning.
Two test scenarios were conducted, corresponding to constant heart rates of \qty{60}{\BPM} and \qty{90}{\BPM}. 
The mechatronic test bench was utilized to replicate a physiological ICP waveform of the respective heart rate with a mean pressure of \qty{15}{\mmHg} and a peak-to-peak amplitude of \qty{1}{\mmHg}~\cite{Qvarlander2010} in vitro. 
In both test scenarios, the objective was to induce pathological ICP waveform patterns that are typically observed in normal pressure hydrocephalus~\cite{Shprecher2008}. 
Specifically, the peak-to-peak ICP amplitude should be doubled while avoiding any increase in mean ICP. 
For reference learning, the BO algorithm used a total of $n = 20$ iterations to tune the delay and magnitude parameters of the reference trajectory.
The parameters for the initial $n_\mathrm{r} = 5$ BO iterations were sampled uniformly from the feasible parameter set $\Theta$.
The delay was allowed to range from $\qty{0}{\s}$ to $\qty{0.2}{\s}$, ensuring actuation starts before the systolic phase of the heartbeat ends.
The magnitude was allowed to range from $\qty{0.2}{\mm}$ to $\qty{1.25}{\mm}$, ensuring mechanical safety constraints and limiting maximum balloon inflation.
During each BO iteration, $n_\mathrm{p}~=~5$ actuation periods were measured after the change in the periodic disturbance estimates fell below the threshold of $\epsilon~=~\qty{0.01}{\mm}$, and the next candidate parameters were computed using the hyperparameters $\lambda~=~1$ and $\beta~=~2$.
While the BO algorithm is implemented on the actuator system's host PC using GPyTorch~\cite{Gardner2018}, the data acquisition and processing is conducted on the MicroLabBox. 
For data transfer between the host PC and the MicroLabBox, serial communication with the RS232 standard is established using the PySerial library and the dSpace RTI library.

The quantitative results for both test scenarios are presented in Table~\ref{tab:actuation}. 
With amplification factors of \qty{1.885}{} in the \qty{60}{\BPM} test scenario and \qty{2.099}{} in the \qty{90}{\BPM} test scenario, the desired doubling of the amplitude was closely achieved. 
The relative increases in the respective mean ICP of \qty{0.55}{\percent} and \qty{1.28}{\percent} are negligibly small, as desired. 
The controller frequency of \qty{100}{\Hz} was thus sufficiently fast to achieve the desired waveform modulation.
For the \qty{90}{\BPM} test scenario, detailed results of the BO algorithm are shown in Figure~\ref{fig:BO}.
During this tuning process, the reference trajectory was iteratively updated and accurately tracked by the MPC, leading to progressively improved ICP modulation. 
Reference tracking and actuation-based ICP waveform modulation corresponding to the first six iterations of the BO algorithm are shown in Figure~\ref{fig:actuation_tuning}.
Improvements are particularly evident when comparing settings 5 and 6 in Figure~\ref{fig:actuation_tuning}, where changes in the learned trajectory directly translated into significant better ICP waveform modulation.
The ICP waveform modulation using the final parameters resulting from the BO algorithm is shown in Figure~\ref{fig:actuation_final}.
Instead of relying on time-consuming and error-prone hand-tuning, the desired ICP waveform modulation was achieved in an automated and safe manner within less than \qty{2}{\min}.
As experimental testing was limited to realistic but constant heart rate scenarios, future research will have to address optimal reference learning for time-varying heart rates, e.g., with BO algorithms that use heart rate as an contextual input.

\begin{table}
    \centering
    \setlength\extrarowheight{2pt}
    \caption{Experimental Waveform Modulation Results}
    \label{tab:actuation}
        \begin{tabularx}{\columnwidth}{lll}
        \toprule
        \textbf{Experiment} & \textbf{Criteria \quad} & \textbf{Results} \\
        \midrule
        \multirow{10}{*}{\textbf{60 BPM}} & Peak-to-peak ICP - baseline & \qty{0.903}{\mmHg} \\
                                            & Peak-to-peak ICP - actuated & \qty{1.704}{\mmHg} \\
                                            & Amplification factor & 1.885 \\
                                            \cmidrule(l){2-3}
                                            & Mean ICP - baseline & \qty{15.197}{\mmHg} \\
                                            & Max absolute mean ICP increase & \qty{0.084}{\mmHg} \\
                                            & Max relative mean ICP increase & \qty{0.550}{\percent} \\
                                            \cmidrule(l){2-3}
                                            & Optimal reference delay & \qty{0.060}{\s} \\
                                            & Optimal reference magnitude & \qty{1.066}{\mm} \\
        \midrule
        \multirow{10}{*}{\textbf{90 BPM}}   & Peak-to-peak ICP - baseline & \qty{0.991}{\mmHg} \\
                                            & Peak-to-peak ICP - actuated & \qty{2.080}{\mmHg} \\
                                            & Amplification factor & 2.099 \\
                                            \cmidrule(l){2-3}
                                            & Mean ICP - baseline & \qty{14.668}{\mmHg} \\
                                            & Max absolute mean ICP increase & \qty{0.188}{\mmHg} \\
                                            & Max relative mean ICP increase & \qty{1.280}{\percent} \\
                                            \cmidrule(l){2-3}
                                            & Optimal reference delay & \qty{0.069}{\s} \\
                                            & Optimal reference magnitude & \qty{0.955}{\mm} \\
        \bottomrule
    \end{tabularx}
\end{table}

\section{CONCLUSION}\label{sec:conclusion}
This work introduces a learning-based model predictive control (MPC) framework to safely achieve a desired intracranial pressure (ICP) waveform modulation with a soft robotic actuator system. 
For accurate and constraint-aware tracking of the system's motor position reference trajectory, the framework combines a tracking MPC with a disturbance observer that exploits the periodic operation of the system.
As the exact influence of the motor position on the ICP cannot be modeled precisely, the framework employs a Bayesian optimization algorithm for online learning of the optimal motor position reference trajectory that yields the desired ICP waveform modulation. 
Experimental validation using a mechatronic test bench replicating physiological ICP waveform patterns demonstrates that the soft robotic actuator system efficiently and safely learns how to double the ICP peak-to-peak amplitude without significantly raising the mean ICP, thereby inducing pathological waveform patterns observed in normal pressure hydrocephalus. 
The presented technical contributions are of high significance in neurological research and clinical practice, as they not only enhance the safety of in vivo interventions and eliminate time-consuming and error-prone hand-tuning efforts, but could also improve experimental outcomes through more precise ICP waveform modulation capabilities.
The introduced learning-based MPC framework may make the soft robotic actuator system a valuable tool for studying cerebrospinal fluid dynamics and related neurological conditions under controlled experimental conditions, as exemplified by the presented application in hydrocephalus research.

\ifthenelse{\boolean{anonymous}}
{
}
{
\section*{ACKNOWLEDGMENT}
We thank Prof. Steffen Leonhardt and Dr. Marian Walter from RWTH Aachen University for providing the test bench.
}

\bibliography{references}
\bibliographystyle{ieeetr}
\end{document}